\documentclass{article}

\usepackage{microtype}
\usepackage{graphicx}
\usepackage{subfigure}
\usepackage{booktabs} 
\usepackage{wrapfig}
\usepackage{csquotes}
\usepackage{hyperref}

\usepackage[accepted]{icml2023}

\usepackage{amsmath}
\usepackage{amssymb}
\usepackage{mathtools}
\usepackage{amsthm}

\usepackage[capitalize,noabbrev]{cleveref}
\usepackage{multirow}

\theoremstyle{plain}

\theoremstyle{definition}

\theoremstyle{remark}

\usepackage[textsize=tiny]{todonotes}

\icmltitlerunning{TaxDiff: Taxonomic-Guided Diffusion Model for Protein Sequence Generation}

\begin{document}

\twocolumn[
\icmltitle{TaxDiff: Taxonomic-Guided Diffusion Model for Protein Sequence Generation}




\icmlsetsymbol{equal}{*}

\begin{icmlauthorlist}
\icmlauthor{Zongying Lin}{pku}
\icmlauthor{Hao Li}{pku,pclab}
\icmlauthor{Liuzhenghao Lv}{pku}
\icmlauthor{Bin Lin}{pku}
\icmlauthor{Junwu Zhang}{pku}
\icmlauthor{Calvin Yu-Chian Chen}{pku}
\icmlauthor{Li Yuan}{pku,pclab}
\icmlauthor{Yonghong Tian}{pku,pclab}
\end{icmlauthorlist}

\icmlaffiliation{pku}{Peking University}
\icmlaffiliation{pclab}{Peng Cheng Laboratory}

\icmlcorrespondingauthor{Li Yuan}{yuanli-ece@pku.edu.cn}
\icmlcorrespondingauthor{Yonghong Tian}{yhtian@pku.edu.cn}

\icmlkeywords{Machine Learning, ICML}

\vskip 0.3in
]



\printAffiliationsAndNotice{}  

\begin{abstract}
Designing protein sequences with specific biological functions and structural stability is crucial in biology and chemistry. 
Generative models already demonstrated their capabilities for reliable protein design. However, previous models are limited to the unconditional generation of protein sequences and lack the controllable generation ability that is vital to biological tasks.
In this work, we propose TaxDiff, a taxonomic-guided diffusion model for controllable protein sequence generation that combines biological species information with the generative capabilities of diffusion models to generate structurally stable proteins within the sequence space. 
Specifically, taxonomic control information is inserted into each layer of the transformer block to achieve fine-grained control. The combination of global and local attention ensures the sequence consistency and structural foldability of taxonomic-specific proteins.
Extensive experiments demonstrate that TaxDiff can consistently achieve better performance on multiple protein sequence generation benchmarks in both taxonomic-guided controllable generation and unconditional generation. Remarkably, the sequences generated by TaxDiff even surpass those produced by direct-structure-generation models in terms of confidence based on predicted structures and require only a quarter of the time of models based on the diffusion model. The code for generating proteins and training new versions of TaxDiff is available at:~\url{https://github.com/Linzy19/TaxDiff}.
\end{abstract}
\section{Introduction}
\label{introduction}

\begin{figure}[ht]
\begin{center}
\centerline{\includegraphics[width=\columnwidth]{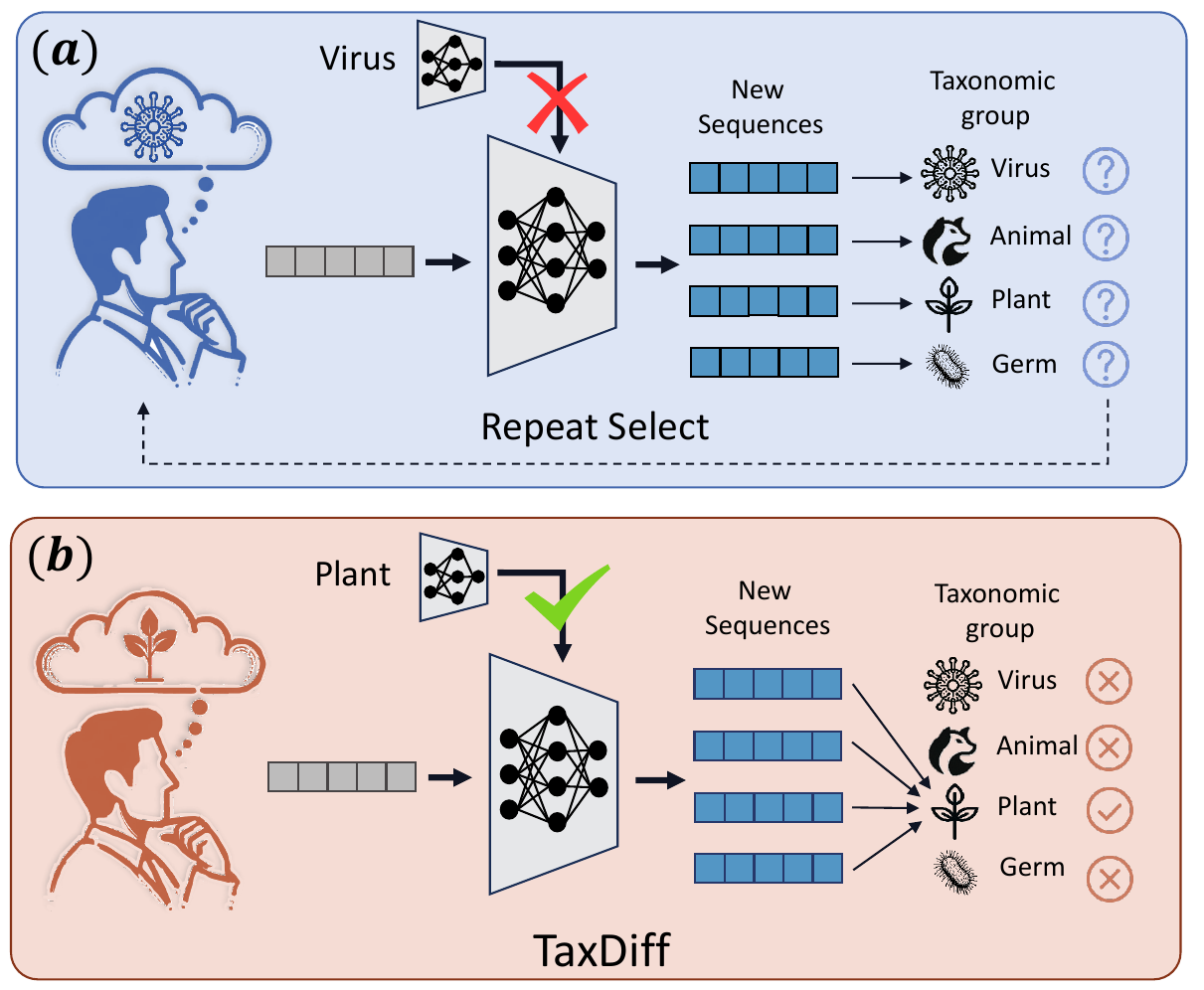}}
\vskip -0.1in
\caption{
\textbf{(a).}~Traditional protein sequence generation models operate without control signals, thus researchers can only randomly generate sequences and subsequently filter them until they fulfill the desired criteria. \textbf{(b).}~Our TaxDiff takes species features as guidance for controllable protein sequence generation, meeting the need of biology downstream tasks.
}
\vskip -0.2in
\label{fig1_idea_map}
\end{center}
\vskip -0.2in
\end{figure}

Protein design~\cite{alphafold1,song2023importance,zheng2023structure} aims to generate protein variants with targeted biological functions, which is significant in multiple biological areas, including enzyme reaction catalysis~\cite{proteinGan,fox2007improving}, vaccine design~\cite{romero2009exploring,lin2022language,phillips2022generating}, and fluorescence intensity~\cite{biswas2021low,tan2023novo}.

Protein design contains two paradigms: sequence generation~\cite{evodiff,VAE_protein_variants,proteinGan} and structure generation~\cite{foldingdiffusion,RFdiffusion}. 
Recently, EvoDiff~\cite{evodiff} proposed a universal designing paradigm, combining structure and sequence generation using the diffusion framework~\cite{score_diffusion}, which improves the protein design efficiency.

Despite the success of EvoDiff~\cite{evodiff} and other sequences generative models~\cite{proteinGan,autogress_antibody_shin,VAE_protein_variants} that are widely used for designing biologically plausible protein sequences, these protein design models are limited to unconditional generation. As shown in Figure~\ref{fig1_idea_map}.(a), in practical scenes, biological researchers need to filter the randomly generated proteins to fulfill the desired criteria~\cite{nature_protein_design}, which is time-consuming and labor-intensive. Thus, unconditional protein generation, which can not control protein properties, is still some way from practical application.

To address the uncontrollable challenge, we propose a taxonomic-guided diffusion model, TaxDiff, to design target proteins with the biological-species control signals.
Specifically, TaxDiff inserts the taxonomic control features into each Denoise Transformer block of the diffusion model to achieve controllable generation. For fine-grained protein sequence generation, we also propose the patchify attention mechanism in the Denoise Transformer block to capture the protein feature on global and local scales. Furthermore, we reclassify protein sequences at the family and species levels to consolidate the overly detailed classification units within UniProt~\cite{uniprot}.
Our TaxDiff follows the protein design paradigm of EvoDiff~\cite{evodiff}. Thus, TaxDiff is capable of generating both protein sequences and structures in a shared space.

We carry out extensive experiments to evaluate TaxDiff across multiple benchmarks, encompassing both unconditional and taxonomic-guided controllable protein sequence generation. In unconditional protein sequence generation, the sequence-based TaxDiff demonstrated comparable structural modeling capabilities to structure-based protein generation models, even significantly outperforming them in common metrics such as TM-score, RMSD, and Fident, with improvements of 11.93\%, 5.4552, and 7.13\% respectively. In taxonomic-guided controllable protein sequence generation, the pLDDT scores from protein structure prediction model OmegaFold~\cite{OmegaFold} far surpassed other sequence generation models, nearing the levels of natural protein sequences. Empirical studies also indicated that due to the patchify attention mechanism, the efficiency of TaxDiff was markedly enhanced, requiring as little as 24 minutes to generate 1000 protein sequences, which is only 1/4 to 2/3 of the time required by other models. All experimental results demonstrate that TaxDiff possesses superior capabilities in exploring protein sequence space and producing structurally coherent proteins. 
The main contributions of our study are outlined as follows:

$\bullet$~To the best of our knowledge, our TaxDiff is the first controllable protein generation model utilizing guidance from taxonomies. 

$\bullet$~Our TaxDiff proposes a taxonomic-guided framework that fits all diffusion-based protein design models. We also propose the patchify attention mechanism for better protein design.


$\bullet$~Experiments demonstrate that our TaxDiff achieves state-of-the-art results in both taxonomic-guided controllable and unconditional protein sequence generation, excelling in structural modeling scores and sequence consistency.

\section{Related Work}

\textbf{Diffusion Models}
Recently, diffusion models~\cite{denoising_diffusion, score_diffusion,jin2023diffusionret} have demonstrated impressive applications and results across various domains, including computer vision~\cite{stable_diffusion}, natural language generation~\cite{NLP_diffusion_discrete}, molecular modeling~\cite{material_Design}, and protein structure generation~\cite{RFdiffusion}.
The inherent suitability of diffusion models for processing protein structures, coupled with the discrete representation of protein sequences, has led to a predominant focus on structural generation in most current protein studies based on diffusion models, with relatively less attention paid to sequence generation.

\textbf{Protein Structures Generation}  
In recent developments, SMCDiff~\cite{SMCDiff}, FrameDiff~\cite{E3}, and Liu~\yrcite{3d_mol_gene} employ the graph neural network to generate functional proteins or molecules. Meanwhile, RFdiffusion~\cite{RFdiffusion} acquires a generative model by fine-tuning the RoseTTAFold structure prediction network~\cite{rosettafold}, achieving notable performance across various fields. FoldingDiff~\cite{foldingdiffusion} describes protein structures through consecutive angles, generating novel structures by denoising from a random, unfolded state.
However, a significant limitation in the structure generation paradigm~\cite{Chroma,E3_generating,ig_VAE,protein_structures,protein_structures2},
equivalent to the sequence data, there is limited data on protein structures, which restricts the potential exploration of the functional space of proteins. 
Meanwhile, sequence-based TaxDiff bypasses the bottleneck of limited training data.

\textbf{Protein Sequences Generation}  
Hawkins-Hooker~\yrcite{VAE_protein_variants} and Repecka~\yrcite{proteinGan} respectively used VAE and GAN to synthesize functionally active enzyme sequences.
Addressing the challenge of nanobodies' complementarity determining region, Shin~\cite{autogress_antibody_shin} employed autoregressive models to solve this problem. Wu~\yrcite{transformer_signal_peptides} and Cao~\yrcite{cao2021fold2seq} adopted the Transformer model to design protein sequences.
Nevertheless, existing sequence-based models predominantly exhibit limitations to unconditional or specific species~\cite{song2023importance,VAE_protein_variants} generation. Such constraints underscore a notable gap in the capacity for conditional generation in these models. Through taxonomic guidance, TaxDiff can generate proteins with the biological-species control signals.

\section{Preliminary}
In this section, we first introduce the problem setting of controllable protein sequence generation in Section~\ref{problem_setting}, then describe the Diffusion Models~(Section~\ref{preknowledge_diffusion}), which is utilized as our main generation framework.

\begin{figure*}[ht]
\begin{center}
\centerline{\includegraphics[width=\textwidth]{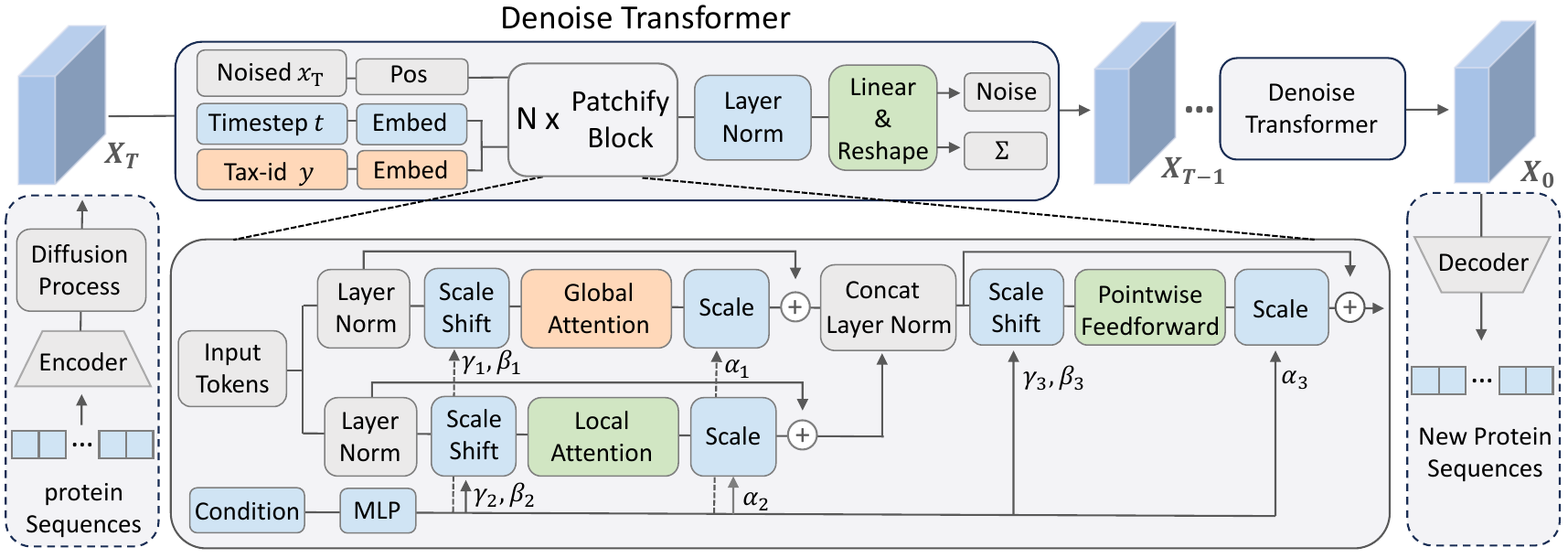}}
\vskip -0.1in
\caption{
\textbf{Overall architecture of the proposed TaxDiff.} This framework delineates how we fuse the Denoise Transformer into the denoising process of the Diffusion model. For a taxonomic-guided controllable generation, we additionally accept a Tax-id $y$ and embed it with Timestep $t$ into the Patchify Blocks. The bottom middle of this framework elaborates on the details in Patchify Block. $\Sigma$ is the predicted diagonal covariance.
}
\label{fig2_framework_map}
\end{center}
\vskip -0.2in
\end{figure*}

\subsection{Protein Sequence Generation}
\label{problem_setting}
In this paper, we consider generating protein sequences under the guidance of taxonomies.
Protein sequence space can be represented as $\mathbf{S} = \langle \mathbf{x}, \mathbf{y} \rangle$ where $\mathbf{x} = (\mathbf{x_1}, \ldots, \mathbf{x_N})\in \mathbf{R}^{N \times L}$ are the protein sequences of length $L$ and \(\mathbf{y} = (\mathbf{y_1}, \ldots, \mathbf{y_M})\in \mathbb{R}^{M \times 1 }\) represents the biological taxonomic category to which the protein belongs, such as Bacteria, Eukaryota, Archaea and so on. $N$ and $M$, respectively represent the total number of protein sequences and categories. We consider the following two generative tasks: 

\textbf{(I) Unconditional generation.} Using the set of proteins $\mathbf{x}$, unconditional generation train parameterized generative models $p_{\theta}(\mathbf{x})$ which can randomly generate diverse and realistic protein sequences without other additional labels.

\textbf{(II) Controllable generation.} With a collection of protein sequences $\mathbf{x}$  with label $\mathbf{y}$, we build a conditional generative model $p_{\theta}( \mathbf{x}|\mathbf{y})$ that is capable of controllable protein sequences generation given desired biological taxonomic category $\mathbf{y}$.

\subsection{Diffusion Model}
\label{preknowledge_diffusion}
We briefly introduce the Diffusion Models~(DMs)~\cite{denoising_diffusion_2015,denoising_diffusion}, which is the generation framework of our TaxDiff. DMs are latent variable models~\cite{podell2023sdxl} that model the data $x_0$ as Markov chains \( x_0 \ldots x_T\), which can be described with two Markovian processes: a forward diffusion process 
$q(x_{1:T} | x_0) = \prod_{t=1}^{T} q(x_t | x_{t-1})$
and a reverse denoising process 
$p_{\theta}(x_{0:T}) = p(x_T) \prod_{t=1}^{T} p_{\theta}(x_{t-1}| x_{t})$. 
The forward process gradually adds Gaussian noise to data \( x_t \):
\begin{equation}
\begin{aligned}
q(x_t|x_{t-1}) = \mathcal{N}(x_t; \sqrt{\bar{\alpha}_t} x_0, (1 - \bar{\alpha}_t)\mathbf{I})
\end{aligned}
\end{equation}
where the hyperparameter $\bar{\alpha}_t$ controls the amount of noise added at each timestep $t$. The $\bar{\alpha}_t$ are chosen such that samples $x_t$ can approximately converge to standard Gaussians $\mathcal{N}(0,\mathbf{I})$. Typically, this forward process $q$ is
predefined without trainable parameters.

The generation process of DMs is defined as learning a parameterized reverse denoising process $p_{\theta}$, which aims to incrementally denoise the noisy variables $x_{T:1}$ to approximate original data $x_0$ in the target data distribution:
\begin{equation}
\begin{aligned}
p_{\theta} (x_{t-1} | x_t) = \mathcal{N}(x_{t-1} ; \mu_{\theta}(x_t, t), \Sigma_{\theta}(x_t, t))
\end{aligned}
\end{equation}
where the initial distribution $p(x_t)$ is defined as $\mathcal{N}(0,\mathbf{I})$.
The means $\mu_{\theta}$ and variances $\Sigma_{\theta}$ typically are neural networks such as U-Nets~\cite{si2023freeu} for images or Transformers for text~\cite{dit}.

\begin{figure*}[ht]
\begin{center}
\centerline{\includegraphics[width=\textwidth]{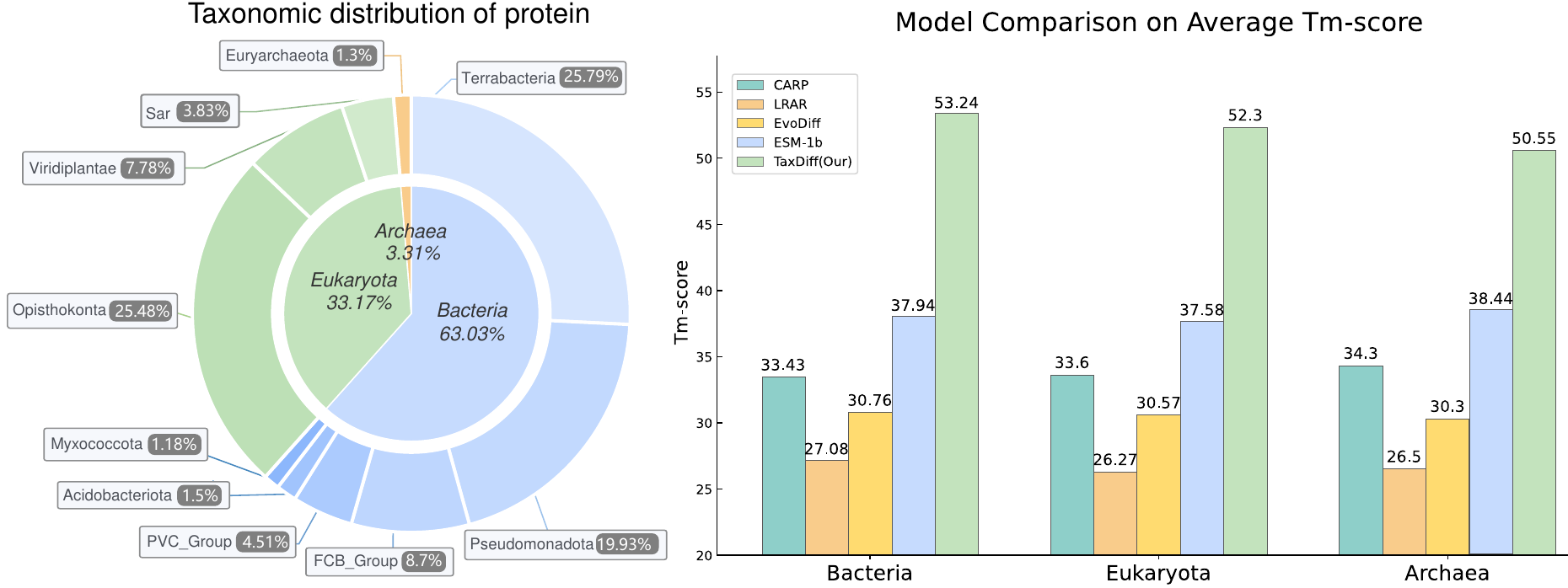}}
\vskip -0.1in
\caption{
\textbf{Taxonomic distribution and comparison of taxonomic-guided models.}
\textit{Left:} We display the distribution of species classification from the second and third taxonomic levels, showcasing the top 10 categories.
\textit{Right:} We compare the performance of different models in terms of TM-score under the condition of the second taxonomic level.
}
\label{fig3_pie_bat_map}
\end{center}
\vskip -0.2in
\end{figure*}

\section{Methodology}
We first introduce the framework and data flow of TaxDiff in Section~\ref{method_1_taxdiff_framework} and then elucidate
the controllable generation under taxonomic-guided in Section~\ref{method_2_conditional_generation}, followed by optimizing the denoise capability of the diffusion model with patchify attention mechanism in Section ~\ref{method_3_patch_attention}. Finally, we will discuss the training procedure, specifically the design of the loss function in Section~\ref{method_4_loss}. The overall architecture of the proposed TaxDiff is illustrated in Figure~\ref{fig2_framework_map}.

\subsection{TaxDiff framework}
\label{method_1_taxdiff_framework}
Recognizing the diversity of amino acids, we have introduced an additional dimension $D$ to enrich features at the amino acid level. Through Encoder, feature-augmented $x$ can thus be represented as \(x \in \mathbb{R}^{L \times D}\). In the Denoise Transformer block, three different types of inputs are processed: the data $x_T$ formed by the forward process in DMs that gradually adds Gaussian noise, the timestep $t$, and the protein taxonomic identifier $y$ (tax-id). $x_T$ undergoes standard Transformer-based frequency position embedding (sine-cosine version)~\cite{vit}, while the timestep $t$ and tax-id $y$ are individually embedded, resulting in two distinct conditional tokens that are concatenated with the $x_T$. Conditional tokens are designed for seamless integration, rendering them indistinguishable from protein sequence tokens. After the terminal Patchify block, these conditional tokens are excised from the sequence. This strategy permits the usage of standard Transformer blocks without modification.

After the patchify block, the sequence tokens must be decoded into a predicted noise and diagonal covariance ($\Sigma$). Both of them retain the shape equivalent to that of the original input. To achieve this, adaptive layer normalization (adaLN)~\cite{adaptive_layer_norm} is applied, and each token is linearly decoded into a tensor of dimensions 
\( L \times 2D\). 
In the decoder, the denoised final result \(\mathbf{x_{0}} \in \mathbb{R}^{L \times D}\) is subjected to an argmax layer: \( argmax(\mathbf{x_{0}}) \in \mathbb{R}^{L}\). The output is then parsed and segmented at each padding or stopping sign, thus generating the protein sequences.

\begin{figure*}[ht]
\begin{center}
\centerline{\includegraphics[width=\textwidth]
{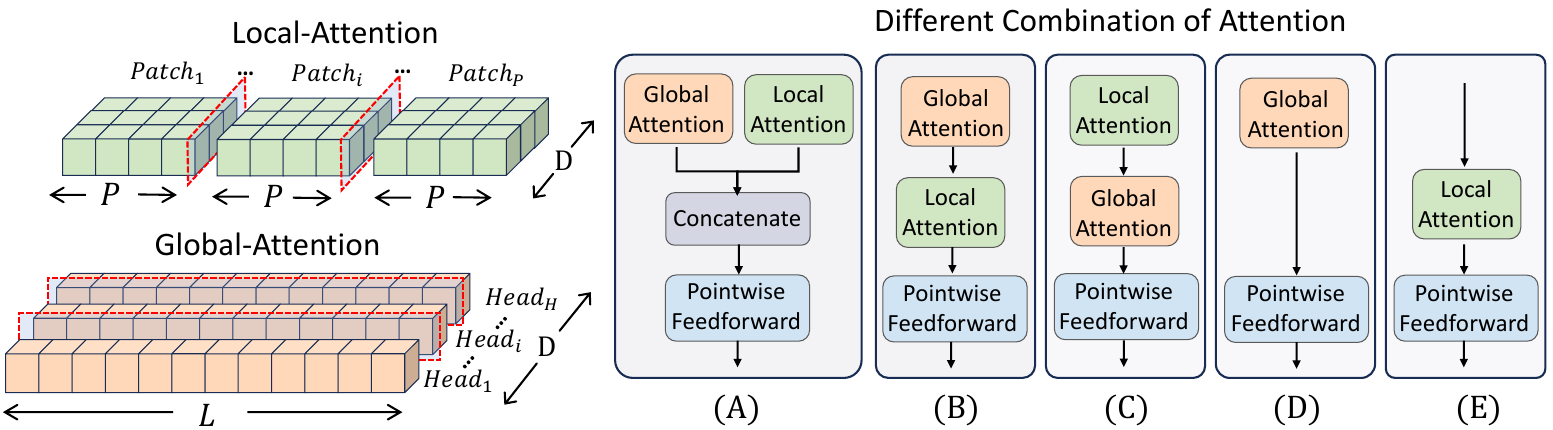}}
\vskip -0.1in
\caption{
\textbf{The patchify attention mechanism.} \textit{Left:} The different division between Local-Attention and Global-Attention. \textit{Right:} we present five different approaches to combining them. The division of the feature dimension in Global-Attention is determined by the number of heads $H$, while the number of tokens created by patchifing is determined by the patchify-size $P$.
}
\label{fig4_reshape_map}
\end{center}
\vskip -0.2in
\end{figure*}

\subsection{Taxonomic-guided generation}
\label{method_2_conditional_generation}
Traditional conditional diffusion models take the class labels or text as extra information~\cite{zhang2023adding,yang2023diffusion}, but our TaxDiff encodes tax-id \( y \) as a condition for controllable generation. In this case, the reverse process is formalized as \( p_\theta(x_{t-1} | x_t, y) \), with both means $\mu_{\theta}$ and variances $\Sigma_{\theta}$ being conditional by \( y \). In this way, the taxonomic-guided encourages the sampling procedure towards maximizing the conditional log-likelihood \( \log p(y | x) \)~\cite{classifier_condition}. Invoking Bayes' Rule, we have \( \log p(y | x) \propto \log p(x | y) - \log p(x) \) and hence \( \nabla_x \log p(y | x) \propto \nabla_x \log p(x | y) - \nabla_x \log p(x) \).
DMs based on score functions guides model sample \( x \) with a heightened probability \( p(x | y) \) by adjusting noise prediction:
\begin{equation}
\begin{aligned}
\hat{\mu}_\theta(x_t, y) = \mu_\theta(x_t, \emptyset) + s \cdot \nabla_x \log p(x | y) \\
\propto
\mu_\theta(x_t, \emptyset) + s \cdot (\mu_\theta(x_t, y) - \mu_\theta(x_t, \emptyset))
\end{aligned}
\end{equation}
where the scalar \( s > 1 \) determines the scale of guidance. Notably, setting \( s = 1 \) reverts the process to the standard sampling. 

As for the DMs with \( y= \emptyset \), it is done by randomly dropping out \( y \) with a certain probability and replacing it with a 'null' embedding \( \emptyset \). This technique of controllable guided yield significantly improved samples over generic sampling techniques~\cite{classifier_condition,guide}. This observed improvement is also replicated in Figure~\ref{fig3_pie_bat_map} and Experiments~\ref{sec_4_3}.

The classification for tax-id $y$ within UniProt is markedly detailed, resulting in the minimal categories that encompass a limited number of a few. These fine-grained categories hinder effective feature extraction across the broader Taxonomic domain. To address this, we reclassify the ninth layer of original taxonomic lineages, which corresponds to the family and species levels~\cite{taonomic_classfication}. 
Specifically, we align the protein sequences from UniProt with the taxonomic data from NCBI~\cite{NCBI} by recursive tracing from the terminal child nodes up to the root. Utilizing the nine layers of classification subordinate to the root node, we assign a novel tax-id $y$ to each sequence. This reorganization effectively condenses the categories to a total of 23,427.
Within this refined classification, cellular organisms represent a predominant 98\% of the protein sequences, with the residual 2\% attributed to viruses and other entries. Predominant domains within cellular organisms include Bacteria (63\%), Eukaryota (33\%), and Archaea (3\%). Further subdivisions and their respective visual representations are elucidated in Figure.~\ref{fig3_pie_bat_map} and Appendix~\ref{append_tax_compare}.

\begin{table*}[htb]
\centering
\footnotesize
\setlength{\tabcolsep}{1.5mm}        
{
\caption{\textbf{Unconditional generation result comparison on AFDB and PDB datasets.} Metrics are calculated with 500 samples generated from each mode. Our TaxDiff best performs on TM-score, RMSD, and Fident for both datasets, and outperforming structure-based RFdiffusion and FoldingDiff on pLDDT.}
\vskip 0.1in
\label{table:1_unconditional_generation}
{
\begin{tabular}{p{50pt}|cc|ccc|ccc}
    \toprule
    \multirow{2}{*}{\textbf{Architecture}}&\multirow{2}{*}{\textbf{Method}}& 
    \multirow{2}{*}{\textbf{pLDDT$\uparrow$}}& \multicolumn{3}{c|}{\textbf{AFDB Dataset}} & \multicolumn{3}{c}{\textbf{PDB Dataset}} \\

    \cmidrule(rl){4-6}\cmidrule(rl){7-9}
     & &  & \textbf{TM-score(\%)$\uparrow$} & \textbf{RMSD$\downarrow$} & \textbf{Fident(\%)$\uparrow$} & TM-\textbf{score(\%)$\uparrow$} & \textbf{RMSD$\downarrow$} & \textbf{Fident(\%)$\uparrow$} \\
    \midrule
    \multirow{2}{*}{CNN} 
    &CARP       & 34.40 ± 14.43 & 25.15 & 18.8463  & 14.23  & 31.21 & 12.5794  & 15.34 \\
    &LRAR       & 49.13 ± 15.50 & 26.44 & 18.0071  & 15.00  & 30.60  & 13.1135  & 15.54 \\
    \midrule
    \multirow{2}{*}{Encoder} 
    &ESM-2       & 51.16 ± 15.52 & 25.06 & 23.6227 & 17.16 & 29.47 & 19.1269  & 17.41\\
    &ESM-1b      & 59.57 ± 15.36 & 33.24 & 20.1126 &  16.94 & 36.85 & 16.5417  & 17.77\\
    \midrule
    \multirow{2}{*}{Decoder}
    & ProtGPT2 & 56.32 ± 16.05 & 25.79 & 19.7728 &  14.48 & 31.11 & 13.2589 &  14.48\\
    & ProGen2 & 61.07 ± 18.45 & 21.59 & 28.1693 &  17.68 & 28.54 & 18.1132 &  17.68\\
    \midrule
    \multirow{2}{*}{Structure}
    &RFdiffusion& 57.27 ± 20.20 & 30.03 & 15.8841  & 13.52 & 32.58 & 12.1724 & 13.76 \\
    &FoldingDiff& 68.89 ± 14.60 & 36.33 & 11.3627  & 19.47 & 37.60  & 9.2003  & 20.05 \\
    \midrule
    \multirow{2}{*}{Diffusion} 
    &EvoDiff    & 44.29 ± 14.51 & 24.22 & 20.0326  & 15.01 & 29.58 & 13.9564   & 15.64\\
    &\textbf{TaxDiff~(Our)}   & \textbf{68.89 ± 9.35} & \textbf{48.26} & \textbf{5.9075} & \textbf{26.60} & \textbf{46.02}  & \textbf{4.5736}  & \textbf{24.13}\\
 \bottomrule
\end{tabular}
}
}
\vskip -0.1in
\end{table*}
\subsection{Patchify attention}
\label{method_3_patch_attention}
To enhance the model's ability to extract global features of protein sequences and simultaneously preserve the salience of amino-acid local features, we have implemented patchify attention mechanism upon the input protein sequences. This procedure is delineated in Figure~\ref{fig4_reshape_map}. 

At the protein sequences level, Global-Attention is employed to capture the intricate relationships between different amino acids. For each head $i$ within the Global-Attention, with a total of $H$ heads, we calculate the \(Q_i = x \times W_i^Q\), \(K_i = x \times W_i^K\) and \(V_i = x \times W_i^V\) as linear transformations of the input matrix \(x \in \mathbb{R}^{L \times D}\), where \( W_i^Q \), \( W_i^K \) and \( W_i^V \) are the weight matrices unique to each head. Attention weights are calculated using the scaled dot-product attention mechanism:
\begin{equation}
\small
    Attention_{\text{head}}(Q_i, K_i, V_i) = softmax \left(\frac{Q_iK_i^T}{\sqrt{d_k}}\right)V_i
\end{equation} Here, \( d_k \) is the dimension of the $K_i$. The final output of the Global-Attention is procured by concatenating the individual heads' outputs and subsequently subjecting them to a linear transformation:
\begin{equation}
\small
    Global\!\!-\!\!Attention(x) = concat(\text{head}_1,  \ldots, \text{head}_H)W^O
\end{equation}
where each \( \text{head}_i \) represents an attention block, and \( W^O \) serves as an output weight matrix to synthesize all Heads.

At the amino-acids level, Local-Attention divides the sequences of length $L$ based on the patchify-size $P$. For each patch \( j \) , the queries ($Q_j$), keys ($K_j$) and values ($V_j$) are deduced by:
\begin{align}
\small
    Q_{j} &= Patch_j(x) \times W_{\text{j}}^Q \\
    K_{j} &= Patch_j(x) \times W_{\text{j}}^K \\
    V_{j} &= Patch_j(x) \times W_{\text{j}}^V
\end{align}
where \(Patch_j(x)\) signifies the partitioning of \(x\) into its corresponding patch, and \(W_{j}^Q\), \(W_{j}^K\) and \(W_{j}^V\) are the patch-specific weight matrices.
Attention weights within each patch are also computed utilizing the scaled dot-product attention:
\begin{equation}
\small
    Attention_{\text{patch}}(Q_{j}, K_{j}, V_{j}) = softmax \left(\frac{Q_{j}K_{j}^T}{\sqrt{d_{k}}}\right)V_{j}
\end{equation} where \( d_{k} \) is the key's dimension within the patch.
The Local-Attention's ultimate output is derived by concatenating the outputs from all patches:
\begin{equation}
\small
    Local\!\!-\!\!Attention(x) = concat(\text{Patch}_{1}, \ldots, \text{Patch}_{P})
\end{equation}
where \( \text{Patch}_{j} \) equating to \(Attention_{\text{patch}}(Q_{j}, K_{j}, V_{j}) \).

Additionally, we derive the scaling parameter $\alpha$ from any residual connections in the Patchify block, while the scaling parameter $\gamma$ and bias parameter $\beta$ are regressed from the conditioning embedding vectors of $t$ and $y$. The adaptive layer normalization~(adaLN) uniformly applies the same function across all tokens.

To effectively combine Global and Local-Attention, we explored five methodologies for their incorporation and subjected them to extensive experiments. The combinatorial strategies are illustrated in the left of Figure~\ref{fig4_reshape_map}: (A) Global-Attention and Local-Attention are used in parallel, with subsequent extracted features concatenation and fusion via Pointwise Feedforward; (B) and (C) Global-Attention and Local-Attention are deploy in a sequential format, capped with Pointwise Feedforward to predict noise and variance; (D) and (E) isolate the utilization to either Global-Attention or Local-Attention exclusively. The comparison of these five methods and the impact of varying patchify size on local feature extraction capabilities are all explored in the Experiments~\ref{sec_4_4}.

\subsection{Training procedure}
\label{method_4_loss}
The training of diffusion models is aimed to learn the reverse process, which is expressed as \( p_\theta(x_{t-1}|x_t) = \mathcal{N}(\mu_\theta(x_t), \Sigma_\theta(x_t)) \), while the Denoise Transformer is used to estimate \( p_\theta \). The model training within the variational lower bound of the log-likelihood of \( x_0 \), with the exclusion of an additional term irrelevant for training, the loss function can be represented to:  
\begin{equation}
\begin{aligned}
     L(\theta) = & -p(x_0|x_1) \\
     & + \sum_t D_{KL}(q^*(x_{t-1}|x_t, x_0) || p_\theta(x_{t-1}|x_t))
\end{aligned}
\end{equation} 
Given that both \( q^* \) and \( p_\theta \) are Gaussian, the Kullback–Leibler divergence (\(D_{KL}\)) can be evaluated by the means and covariances of these distributions.

For simple model training, \( \mu_\theta \) is reparameterized as a noise prediction network \( \epsilon_\theta \), then the model can be trained using a simple mean-squared error function (MSE loss) between the predicted noise \( \epsilon_\theta(x_t) \) and the actual sampled Gaussian noise \( \epsilon_t \): 
\begin{equation}
    L_{MSE}(\theta) = ||\epsilon_\theta(x_t) - \epsilon_t||^2 
\end{equation} 
However, for training diffusion models with a learned reverse process co-variance \( \Sigma_\theta \), it becomes imperative to optimize the entire \(D_{KL}\) term. Followed DiT~\cite{dit}, we  train \( \epsilon_\theta \) with \( L_{MSE} \) and \( \Sigma_\theta \) with the full loss function \(L(\theta)\). Upon the successful training of \( p_\theta \), new protein sequences can be generated by initializing \( x_{t} \sim \mathcal{N}(0, I) \) and subsequently sampling \( x_{t-1} \sim p_\theta(x_{t-1}|x_t) \) via the reparameterization trick.

\begin{table*}[htb]
\centering
\footnotesize
\setlength{\tabcolsep}{1.5mm}  
{
\caption{\textbf{Controllable generation on AFDB and PDB datasets.} Metrics are calculated with 1000 samples generated from each model, with lengths following a random distribution between 10 and 256. The sampling time was recorded on a single 4090 for 1000 samples.}
\label{table:2_conditional_generation}
\vskip 0.1in
{
\begin{tabular}{llc|ccc|ccc}
\toprule
\multirow{2}{*}{\textbf{Method}}&\multirow{2}{*}{\textbf{pLDDT}}& 
\multirow{2}{*}{\textbf{Time(mins)$\downarrow$}}& \multicolumn{3}{c|}{\textbf{AFDB Dataset}} & \multicolumn{3}{c}{\textbf{PDB Dataset}} \\

\cmidrule(rl){4-6}\cmidrule(rl){7-9}
 &  &  &\textbf{TM-score(\%)$\uparrow$} & \textbf{RMSD$\downarrow$}  & \textbf{Fident(\%)$\uparrow$} & \textbf{TM-score(\%)$\uparrow$} & \textbf{RMSD$\downarrow$}  & \textbf{Fident(\%) $\uparrow$} \\
\midrule
CARP & 46.84 ± 13.35 & 96.6 & 33.62 & 12.5753  & 12.5  & 32.38 & 10.8296  & 11.82  \\
LRAR & 47.33 ± 14.26 & 79.31 & 26.94 & 17.9833  & 15.83 & 30.41 & 13.9929  & 16.33 \\
Evodiff & 56.28 ± 16.52  & 99.75 & 30.22 & 15.8664  & 16.17 & 33.02 & 12.3685  & 16.29\\
ESM-1b & 67.91 ± 11.59 & 37.4 & 37.13 & 13.6791  & 16.76 & 41.90 & 9.8445  & 17.27\\
\textbf{Taxdiff(Our)} & \textbf{69.00 ± 9.13} &\textbf{24.53} & \textbf{49.27} & \textbf{5.6518} & \textbf{25.02} & \textbf{48.80} & \textbf{4.8453}  & \textbf{24.85}  \\
\bottomrule
\end{tabular}
}
}
\vskip -0.1in
\end{table*}
\section{Experiments}
\label{experiments}
In this section, we justify the advantages of TaxDiff through comprehensive experiments. We begin by presenting our experimental setup in Section~\ref{sec_4_1}. Subsequently, we report and analyze the outcomes of unconditional and controllable generation in Section~\ref{sec_4_2} and Section~\ref{sec_4_3}, respectively. Finally, We explore various methods of combining Global and Local-Attention, and examine the impact of different patchify-size in Section~\ref{sec_4_4}.

\subsection{Experiment Setup}
\label{sec_4_1}
\textbf{Datasets:} For our model training, we utilized the Uniref50 dataset from Uniprot~\cite{uniprot}, a protein database formed through clustering. Within TaxDiff, we filter protein sequences in Uniref50 that were less than 256 amino acids in length. Sequences falling short of 256 amino acids undergo zero-padding to equalize their length, thereby standardizing the representation of proteins as sequences with a uniform length \( L = 256 \). 

\textbf{Setting:} Following the DiT~\cite{dit}, we replace the standard layer norm in the Transformer block with adaLN and initialize batch normalization to zero within each Patchify block (adaLN-zero)~\cite{zero_init} and employed AdamW~\cite{AdamW} for training all models. A constant learning rate of $1 \times 10^{-4}$ without weight decay is used for training, while the batch size is set to 512 on eight 4090 GPUs. Results in~\cref{table:1_unconditional_generation} and~\cref{table:2_conditional_generation} have the same setup with patchify-size $P=16$, the layer of patchify block $N = 12$. More details about setting in Appendix~\ref{append_experiment_setting}.

\textbf{Evaluation:} We measure model performance through the foldability and consistency of sequences generated by TaxDiff.
We use Omegafold~\cite{OmegaFold}, which relies only on a single sequence without homologous or evolutionary information, to give the confidence scores(pLDDT) in structure prediction for each residue. Subsequently, Foldseek~\cite{foldseek} calculates the average TM-score,  which reflects the structural similarity, and the average RMSD, which focuses on the protein structure size and local variations. Additionally, Foldseek provides the Fident, which serves as sequence-level homology. We choose natural protein structures from the Protein Data Bank (PDB)~\cite{pdb} to verify the natural-like degree of generated sequences and high-confidence protein structures predicted by Alphafold (AFDB)~\cite{alphafold1,alphafold2} to expand the comparison range and verify the broad validity of our model. Settings for results in~\cref{table:1_unconditional_generation} and~\cref{table:2_conditional_generation} are same. More Details are provided in the Appendix~\ref{append_evalution}

\textbf{Baselines:}
We compare TaxDiff to several competitive baseline models.
The left-to-right auto regressive~(LRAR) and convolutional autoencoding representations of proteins(CARP)~\cite{Carp} are both trained with the same dilated convolutional neural network architectures on the UniRef50.
FoldingDiff~\cite{foldingdiffusion} and RFdiffusion~\cite{RFdiffusion} are
recent progress on diffusion models for protein structure generation. 
Notably, the RFdiffusion and FoldingDiff directly produce protein structures, so we first unconditionally select structures generated by these two and use ESM-IF to design their corresponding sequences. 
EvoDiff~\cite{evodiff} is a diffusion model combining evolutionary-scale data based on programmable, sequence-first design.
ESM-1b~\cite{esm1b} and state-of-the-art ESM-2~\cite{esm2} are the protein masked language models, which were trained on different releases of UniRef50. 
ESM series both generated many “unknown” amino acids, and performance was improved results by manually setting the logits for X to inf.
ProtGPT2~\cite{protgpt2} and ProGen2~\cite{progen2} are autoregressive large protein language models based on GPT2~\cite{gpt2}, which have undergone pre-training in UniRef50 and UniRef90 respectively.

\begin{table*}[htb]
\centering
\footnotesize
\setlength{\tabcolsep}{2.2mm}  
{
\caption{\textbf{Contrast experiment of different patchify-size $P$ in Local-Attention} on AFDB and PDB datasets. Metrics are calculated with 1000 samples generated from each model.}
\label{table:4_patch_size}
\vskip 0.1in
{
\begin{tabular}{clc|ccc|ccc}
\toprule
 \multirow{2}{*}{\textbf{Patchify-size}} 
 & \multirow{2}{*}{\textbf{pLDDT$\uparrow$}} & \multirow{2}{*}{\textbf{Time(mins)$\downarrow$}}
 & \multicolumn{3}{c|}{\textbf{AFDB Dataset}} & \multicolumn{3}{c}{\textbf{PDB Dataset}} \\
 \cmidrule(rl){4-6}\cmidrule(rl){7-9}
  & & &  \textbf{TM-score$\uparrow$} & \textbf{RMSD$\downarrow$} & \textbf{Fident(\%)$\uparrow$} & \textbf{TM-score$\uparrow$} & \textbf{RMSD$\downarrow$} &\textbf{ Fident(\%)$\uparrow$} \\
\midrule
4  & 67.56 ± 10.35 & \textbf{20.57} & 46.70 & 6.8047 & 23.01 & 46.58 & 6.2253 & 20.57  \\
8  & 68.88 ± 9.55 & 22.94 & 45.57 & 6.5407 & 23.94 & 46.04 & 5.4404 & 22.94  \\
16 & 69.01 ± 9.03 & 24.85 & \textbf{48.88} & \textbf{5.4992} & 25.88 & \textbf{49.58} & \textbf{4.6262} & 24.95  \\
32 & 65.25 ± 12.12 & 20.72 & 42.82 & 8.094 & 21.71 & 43.27 & 6.8687 & 20.72  \\
64 & \textbf{70.83 ± 8.77} & 24.78 & 46.48 & 5.9851 & \textbf{26.16} & 45.88 & 5.3077 & \textbf{25.20} \\
\bottomrule
\end{tabular}
}
\vskip -0.1in
}
\end{table*}

\subsection{Unconditional Sequences Generation }
\label{sec_4_2}
In the unconditional generation, TaxDiff sets the condition $y$ to be \( \emptyset \) and generates 500 protein sequences within the length range of $10$-$256$. For sequences lengths shorter than 10, we consider them to be invalid protein sequences and, therefore remove them, which is a practice also applicable to sequences generated by all other models. 

As shown in Table~\ref{table:1_unconditional_generation}, TaxDiff outperforms competitive baseline methods across all metrics with a noticeable margin. It's important to note that the sequences generated by our model have exceeded those of the structure generation model RFdiffusion and FoldingDiff in terms of pLDDT. This underscores our model's efficacy in extracting structural modeling information from protein sequences and substantiates the effectiveness of our sequence-based modeling paradigm. Moreover, in the structural alignment with AFDB and PDB datasets, TaxDiff significantly improves TM-score and RMSD, substantially outperforming other models, especially RMSD, which is less than half that of other models. Furthermore, the sequence consistency Fident also surpasses other models on both two datasets, showcasing the comprehensiveness and generalization capability of our model. Overall, the superior performance demonstrates TaxDiff's enhanced ability to simulate protein sequence distributions and generate authentic and highly consistent protein sequences.

\subsection{Taxonomic-Guided Sequences Generation}
\label{sec_4_3}
Under the taxonomic-guided, our objective is to perform controllable protein sequence generation to meet specific needs for proteins of particular species. This can be useful in realistic settings of protein and drug design where we are interested in discovering proteins with specific taxonomic preferences. 
For a valid comparison, we fine-tune a subset of representative networks with taxonomic-guided and same-label embedding architecture, enabling them to learn the distribution of taxonomic categories. Then, We set the conditional tax-id $y$ to a fixed random variable representing the taxonomic group to which the protein belongs and generate 1,000 protein sequences to test the model's controllable generation ability. 

We report the numerical results in Table~\ref{table:2_conditional_generation}. As shown in the table, TaxDiff significantly outperforms baseline models on all the metrics, including the previous diffusion model named EvoDiff, running on the whole Uniref50. The results
demonstrate that by taxonomic-guided, not only TaxDiff, almost all models acquired a higher capacity to incorporate given taxonomic information into the generation process. Furthermore, we compare the time to generate 1000 protein sequences with these models. TaxDiff, fusing transformer into diffusion architectures, not only requires less inference time than models relying solely on either the Transformer architecture like ESM-1b or solely on the Diffusion architecture like Evodiff but also significantly outperforms other models in all metrics.

\subsection{Sequences Global and Local Attention}
\label{sec_4_4}
In this section, we investigate the impact of different attention combinations on model performance and further analyze the effect of patchify-size at the amino acid level. Specifically, we examine the influence of five different attention combinations in Figure~\ref{fig4_reshape_map} on the protein sequence generation capability. To standardize comparison, except for the experimental variables, we keep other parameters consistent and fix the random seed. This allows the five attention combinations to generate corresponding protein sequences under 1,000 fixed random taxonomic groups. The evaluation of the generated sequences is the same as controllable sequence generation.

The visual comparison in Figure~\ref{fig5_ablation_v} and experimental results in Appendix~\ref{table:3_combine_compare} indicate that the parallel use of Global-Attention and Local-Attention (Method A) achieves the best performance across all metrics, significantly surpassing other combinations. However, it is noteworthy that the exclusive use of Local-Attention (Method E) secures the second-best result, exceeding the performance of what we initially anticipated as the sub-optimal (method B) and (method C). This suggests that for protein sequences, unlike natural language, focusing on local effects using Local-Attention is more effective than applying attention across the entire sequence. It implies that protein sequences also contain \enquote{short sentence} like local semantic structures.

To further analyze the impact of different patchify-size on features representation at the amino acid level, we divide sequences by length into segments from 4 to 64, keeping other parameters constant, and use Method A to combine Global-Attention and Local-Attention. We report the numerical results in Table~\ref{table:4_patch_size}. The results demonstrate that dividing protein sequences into 16 amino acids per local patch provides significant advantages for protein structure modeling metrics, such as TM-score and RMSD. In contrast, using a larger patchify size, like dividing protein sequences into 64 amino acids per local patch, has a more substantial impact on improving sequence-related metrics like Fident.

\begin{figure}[htb]
\centerline{\includegraphics[width=\columnwidth]{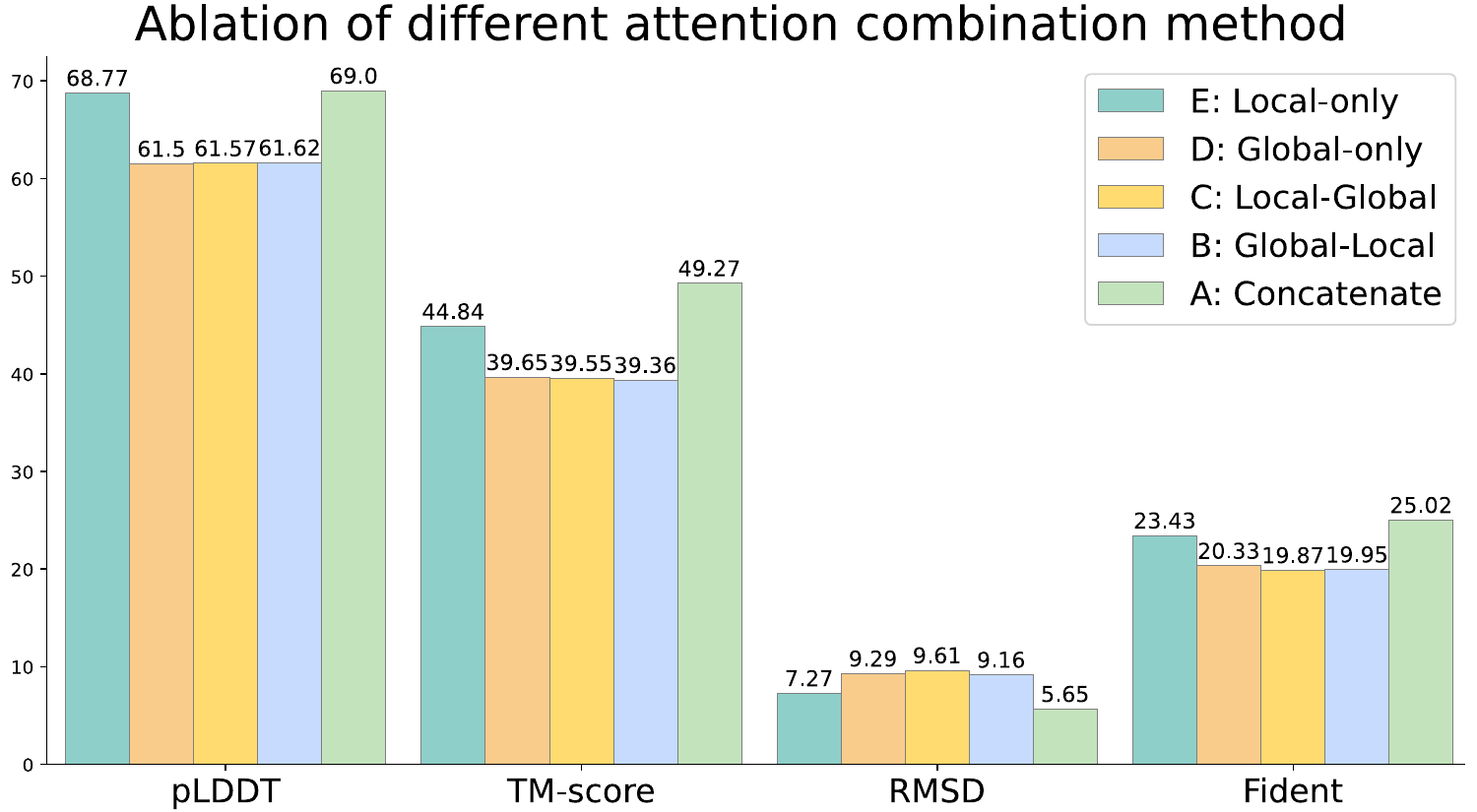}}
\vskip -0.1in
\caption{\textbf{Ablation result of different attention combination} on AFDB datasets. Metrics are calculated with 1000 samples generated from each model.}
\label{fig5_ablation_v}
\vskip -0.2in
\end{figure}
\section{Conclusion and Future Work}
While current models can only perform unconditional sequence generation, TaxDiff overcomes this limitation by learning taxonomic-guided over protein sequences space.
By combining the Global-Attention to sequences and Local-Attention to amino acids, TaxDiff can effectively generate sequences that are structurally reliable and consistent in sequence. Furthermore, TaxDiff requires only a quarter of the time needed by other diffusion-based models  to achieve superior performance. Various experimental results demonstrate its significantly better ability for modeling natural protein sequences. 
As a versatile and principled framework, TaxDiff can be expanded for various protein generation applications in future work. For instance, scaling up  TaxDiff to the more challenging protein generation of protein complex.

\section*{Impact Statements}
Our work with biological protein sequences aims to advance the field of machine learning and introduce machine learning into traditional scientific neighborhoods as a way to promote and robust the AI for science community and accelerate research in the basic sciences. 

Our model TaxDiff, a taxonomy-guided protein sequence generation model can greatly reduce screening time for biological researchers and facilitate drug and new protein design. 

At the same time, however, our work is based on generative models that learn the feature space of protein sequences and use it to generate new protein sequences that may not be easily expressed and synthesized in the lab.

\nocite{langley00}

\bibliography{example_paper}
\bibliographystyle{icml2023}

\newpage
\appendix
\onecolumn
\section{Appendix.}

\subsection{Result and visualization}
\label{append_tax_compare}

Taxonomic distribution and comparison of taxonomic-guided models.
More details of refined classification results will be available in supplementary material and website ~\url{https://github.com/Linzy19/TaxDiff}.
\begin{figure*}[ht]
\vskip 0.2in
\begin{center}
\centerline{\includegraphics[width=\textwidth]{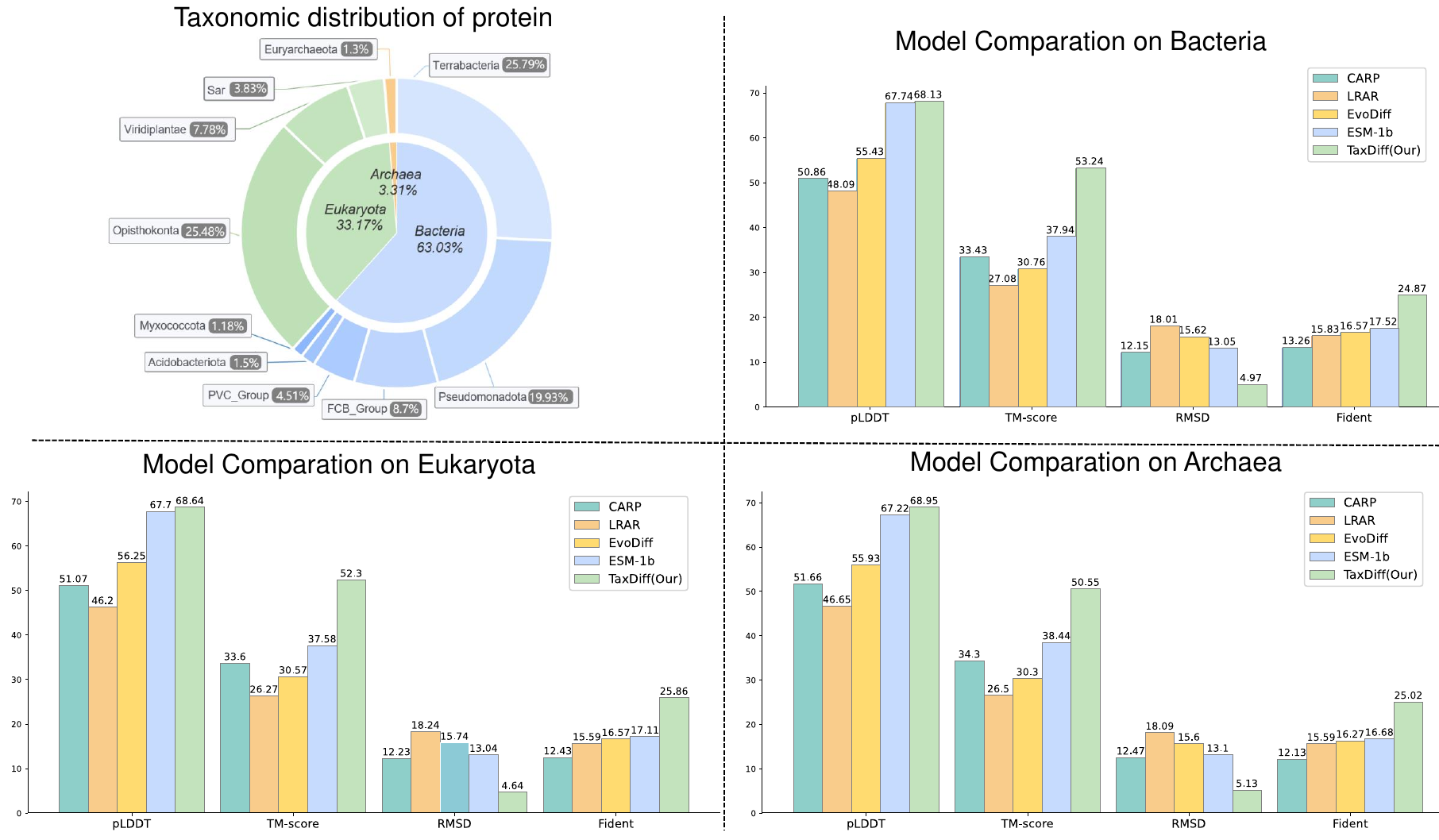}}
\caption{
\textbf{Taxonomic distribution and comparison of taxonomic-guided models.}
\textit{Top Left:} We display the distribution of species classification from the second and third taxonomic levels, showcasing the top 10 categories.
In the bar graph, we compare the performance of different models under the condition of the second taxonomic level.
}
\label{fig6_detail_pie_bat_map}
\end{center}
\vskip -0.2in
\end{figure*}

\subsection{Experiment setting}
\label{append_experiment_setting}
Following common practices in generative modeling literature, we maintained an exponential moving average (EMA) of the TaxDiff weights during training, with a decay factor of 0.9999. The same training hyperparameters were applied across all TaxDiff models and patch sizes. All reported results are based on the EMA model. The same training hyperparameters were applied across all TaxDiff model sizes and patch sizes, almost entirely retained from ADM, without adjustments in learning rate, decay/warm-up schedule, Adam $\beta_1, \beta_2, \beta_3$, or weight decay.

\subsection{Evaluation}
\label{append_evalution}
We measure model performance through sequence consistency and structural foldability, which indicates whether the model can learn the rules of protein sequences from the data. 
For assessing the feasibility of the sequences, We employed OmegaFold~\cite{OmegaFold} to predict their corresponding structures and calculate the average predicted Local Distance Difference Test (pLDDT) across the entire structure, which reflects OmegaFold's confidence in its structure prediction for each residue on sequences level. Omegafold performs structure prediction without the need for homologous sequences or evolutionary information, relying solely on a single sequence for prediction. 
However, due to inherent noise and errors in OmegaFold's structure predictions, which only consider the foldability of individual sequences, we further measure our results using Foldseek~\citep{foldseek}. Foldseek facilitates the pairing of the queried protein \( p^{query} \) with structurally similar proteins from an existing protein database, yielding pairs represented as \((p^{query},\ p^{target})\). Here, \( p^{target} \) denotes the protein in the database with a significant structural similarity to \( p^{query} \). The magnitude of the average template modeling score (TM-score~\citep{Tm-score}) value and root-mean-square deviation (RMSD) reflects the degree of structural similarity. TM-score takes into account the overall topological structure of proteins, focusing more on the protein's overall structure. RMSD calculates the square root of the average position deviation of corresponding atoms between two protein structures, being highly sensitive to the size of the protein structure and local variations. Additionally, Foldseek also calculates the sequence identity (FIDENT) between \( p^{query} \) and \( p^{target} \), reflecting their sequence-level similarity.
In Foldseek, we choose natural protein structures from the Protein Data Bank (PDB)~\cite{pdb} to verify the natural-like degree of our sequences and high-confidence protein structures predicted by Alphafold (AFDB)~\cite{alphafold1} to expand the comparison range and verify the broad validity of our model.

\subsection{Ablation result}
\label{append_ablation}
\begin{table*}[htb]
\centering
\footnotesize
\setlength{\tabcolsep}{2.2mm} 
{
\caption{\textbf{Ablation result of different attention combination} on AFDB and PDB datasets. Metrics are calculated with 1000 samples generated from each model.}
\label{table:3_combine_compare}
{
\begin{tabular}{cl|ccc|ccc}
\toprule
 \multirow{2}{*}{\textbf{Method}} &  \multirow{2}{*}{\textbf{pLDDT$\uparrow$}} & \multicolumn{3}{c|}{\textbf{AFDB Dataset}} & \multicolumn{3}{c}{\textbf{PDB Dataset}} \\
 \cmidrule(rl){3-5}\cmidrule(rl){6-8}
 &  & \textbf{TM-score$\uparrow$} & \textbf{RMSD$\downarrow$} & \textbf{Fident(\%)$\uparrow$} & \textbf{TM-score$\uparrow$} & \textbf{RMSD$\downarrow$} & \textbf{Fident(\%)$\uparrow$} \\
\midrule
E & 68.77 ± 10.67 & 44.84 & 7.2685 & 23.43 & 44.00 & 5.9162 & 24.1\\
D & 61.50 ± 13.55 & 39.65 & 9.2808 & 20.33 & 39.30 & 7.6774 & 19.53\\
C & 61.57 ± 12.99 & 39.55 & 9.6091 & 19.87 & 40.21 & 7.6458 & 20.12\\
B & 61.62 ± 12.99 & 39.36 & 9.1560 & 19.95 & 40.52 & 7.551 &19.77\\
A & \textbf{69.01 ± 9.03} & \textbf{48.88} & \textbf{5.4992} & \textbf{25.88} & \textbf{49.58} & \textbf{4.6262} & \textbf{24.95} \\
\bottomrule
\end{tabular}
}
}
\end{table*}

\end{document}